# Fast Acquisition for Quantitative MRI Maps: Sparse Recovery from Non-linear Measurements

Anupriya Gogna[*] and Angshul Majumdar[+]

[*]IIIT Delhi
Okhla Phase 3
New Delhi, 110020, India
anupriyag@iiitd.ac.in

[+]IIIT Delhi
Okhla Phase 3
New Delhi, 110020, India
angshul@iiitd.ac.in

*Abstract: This work addresses the problem of estimating proton density and T1 maps from two partially sampled K-space scans such that the total acquisition time remains approximately the same as a single scan. Existing multi-parametric non-linear curve fitting techniques require a large number (8 or more) of echoes to estimate the maps – resulting in prolonged (clinically infeasible) acquisition times. Our simulation results show that our method yields very accurate and robust results from only two partially sampled scans (total scan time being the same as a single echo MRI). We model PD and T1 maps to be sparse in some transform domain. The PD map is recovered via standard Compressed Sensing based recovery technique. Estimating the T1 map requires solving an analysis prior sparse recovery problem from non-linear measurements, since the relationship between T1 values and intensity values / K-space samples is not linear. For the first time in this work, we propose an algorithm for analysis prior sparse recovery for non-linear measurements. We have compared our approach with the only existing technique based on matrix factorization from non-linear measurements; our method yields considerably superior results.*

## 1. Introduction

It is well known that measurement of the change in T1 provides important information about the mobility and chemical environment of the tissues of interest. Previous studies examining the utility of T1 mapping in the context of neurologic and psychiatric disease demonstrated variations in T1 values, sometimes very subtle, in specific brain regions within conditions of autism [1], schizophrenia [2], multiple sclerosis [3], tumors [4] and a host of other disorders. The importance of T1 map estimation is appreciated by the MRI community. However, it is not widely used in clinical diagnosis. This owes to the fact that the data acquisition associated with computing the T1 maps is time consuming; it is an order of magnitude higher than that of an ordinary MRI.

The prolonged acquisition time is the main deterrent behind widespread application of quantitative maps in clinical studies. Parallel imaging can reduce the acquisition time around 2/3 fold. In recent times, Compressed Sensing (CS) techniques have been proposed for reducing the acquisition time by partial sampling of the K-space [5-8]; CS techniques can also reduce the scan time by 2/3 fold; however CS based recovery yields good images but poor quantitative map estimates. Hence combining CS with parallel MRI is not viable option for quantitative map estimation. This means that the total data acquisition time required for computing the T1 maps is still significantly higher (4-5 times) than that of an ordinary MRI scan.

In this work we propose a technique to reduce the scan time to the same order of magnitude as an ordinary single echo MRI scan. Recently [9] a matrix factorization based method was proposed to address this problem. It modeled the quantitative map as a low-rank matrix. It required only two echoes to recover the map. In this work we propose to reduce the scan time even further - we want to recover the PD and T1 maps from partially sampled K-space scans such that the total data acquisition time does not exceed that of an ordinary scan. We assume PD and T1 maps to be sparse in some transform domain. There are two advantages of our proposed approach over the prior study [9]:

1. In MRI reconstruction the sparse model is more well known and well understood compared to the low-rank model.

2. Using the sparse model we can reduce the acquisition time even further; we will still need two echoes but the K-space for each echo can be partially sampled effectively leading to the same acquisition time as an ordinary MRI scan.

Studies in CS based MRI reconstruction [10-12] recovered the image from the K-space measurements. The relationship between the measurement (Fourier space) and the image domain is linear. Hence the aforementioned studies recovered a sparse set of coefficients (transform domain sparsity) from linear (Fourier) measurements. In the current scenario, the relationship between the K-space and the T1 map is not linear anymore; therefore in order to recover the T1 map we would need to recover a sparse set of transform coefficients from non-linear measurements. This problem has not been solved before. In this work we derive an efficient algorithm to solve the same.

## 2. T1 Map Estimation

The MR signal intensity is governed by the following equation,

$$x = \rho(1 - e^{-TR/T1})e^{-TE/T2} \qquad (1)$$

where $\rho$ is the proton density (spin density), $TR$ is the repetition time and $TE$ the echo time of the applied magnetization.

For T1 weighting, the echo time is kept small so that the effect of T2 weighting is virtually nullified. This leads to the following approximate relation,

$$x = \rho(1 - e^{-TR/T1}) \qquad (2)$$

In traditional quantitative estimation, the values of $TR$ are varied to acquire multiple echoes (x's) from which the proton density ($\rho$) and the T1 map are estimated using non-linear curve fitting. Acquiring multiple echoes is time consuming. Prior studies used CS based techniques to partially sample the K-space (for each echo); this reduced the acquisition time by 3 or 4 folds. Unfortunately the CS based techniques yielded wonderful image reconstruction but below par estimates of the actual maps [7].

In this work we propose to reduce the scan time for T1 imaging so that it takes the same acquisition time as that of an ordinary single echo MRI scan.

The K-space data acquisition is expressed as:

$$y = Fx + \eta = F\rho(1 - e^{-TR/T1}) + \eta, \ \eta \sim N(0, \sigma^2) \qquad (3)$$

where $F$ is the Fourier transform and $\eta$ is the system noise.

There are two variables (proton density and T1 map) therefore we will require two scans at least. We propose a simple protocol for the scan. From the first scan we need to estimate the proton density (PD); the repetition time is kept high (>5T1) so that the effect of T1 weighting is virtually nullified. Under this condition (3) can be approximated as:

$$y = F\rho + \eta \tag{4}$$

The proton density can be easily recovered by applying the inverse Fourier transform. However, in this case we want to drive the scan time lower, hence we propose to sub-sample the K-space; the data acquisition model is expressed as:

$$y = RF\rho + \eta \tag{5}$$

where $R$ denotes the sub-sampling mask.

The PD can be recovered using standard CS techniques like $l_1$-minimization.

$$\min_\alpha \|y - RF\rho\|_2^2 + \lambda \|\Psi\rho\|_1 \tag{6}$$

Here it is assumed that $\rho$ is sparse in $\Psi$. This is an analysis prior problem. Efficient algorithms for solving this have been proposed in [12].

Once the proton density is recovered, the remaining task is to obtain the T2 map. This is the challenging step. The map is recovered from the second scan. In the second scan the echo time is kept to be of the same order as the T1 values.

In this work, we assume that the T1 map is sparse in a transform domain like wavelet or finite difference. Sparsifying transforms like wavelets encode the discontinuities in the image. If we invert and scale each component of the matrix (TR/T1) the positions of the discontinuities do not change. As T1 values are never 0, the inversion does not introduce any discontinuity either. Thus TE/T2 will also turn out to be sparse in wavelet domain. Replacing Z=TR/T1 in (3) we get,

$$y = F\rho(1 - e^{-Z}) + \eta \tag{7}$$

To reduce the scan time even further we sub-sample the K-space. This leads to:

$$y = RF\rho(1 - e^{-Z}) + \eta \tag{8}$$

The objective is to recover Z (recovering T1 from Z is trivial). The recovery is formulated as:

$$\|y - RF\rho(1 - e^{-Z})\|_2^2 + \lambda \|\Psi z\|_1 \tag{9}$$

This is an analysis prior sparse recovery problem from non-linear measurements. There is no algorithm to solve (9). In the next section we derive a simple algorithm for the same.

## 3. Analysis Prior Sparse Recovery from Non-linear Measurements

### 3.1. Solving Synthesis and Analysis Prior Sparse Recovery from Linear Measurements

First we will briefly review the simplest techniques for sparse recovery (both synthesis and analysis prior). We want to solve:

$$y = Ax + \eta \tag{10}$$

For the synthesis prior problem, x is the sparse signal to be recovered, A is the linear measurement matrix. It (10) is solved via $l_1$-minimization.

$$\min_x \|y - Ax\|_2^2 + \lambda \|x\|_1 \tag{11}$$

Majorization of (11) leads to [15]:

$$\min_x \|b - x\|_2^2 + \frac{\lambda}{a} \|x\|_1 \tag{12}$$

where $b = x_k + \frac{1}{a} A^T (y - Ax_k)$ at the $k^{th}$ iteration.

Here $a$ is the maximum eigenvalue of $A^T A$.

The closed form update for (12) is iterative soft thresholding [16].

$$x_{k+1} = signum(b) \max(0, |b| - \frac{\lambda}{2a}) \tag{13}$$

For the analysis prior it is assumed that the signal is not sparse, but is sparse in a transform domain ($\Psi$). The corresponding optimization problem is:

$$\min_x \|y - Ax\|_2^2 + \lambda \|\Psi x\|_1 \tag{14}$$

The first step to solve (14) is the same as the synthesis prior. Majorization leads to:

$$\min_x \|b - x\|_2^2 + \frac{\lambda}{a} \|\Psi x\|_1 \tag{15}$$

The updates for solving (15) are given in [12]; they are:

$$z_k = (\frac{2a}{\lambda} D^{-1} + cI)^{-1} (cz_{k-1} + \Psi(b - \Psi^T z_{k-1})) \tag{16a}$$

where $D = diag(|\Psi b|^{-1})$

$$x_k = b - \Psi^T z_k \tag{16b}$$

## 3.2. Solving Synthesis Prior Sparse Recovery from Non-Linear Measurements

There have been some prior studies on solving the synthesis prior sparse recovery problem from non-linear measurements [13, 14]; this was based on modifying the ISTA (12) and (13). There are two steps in ISTA –

$$b = x_k + \sigma A^T (y - Ax_k); \quad \sigma = 1/a$$

$$x = signum(b) \max(0, |b| - \frac{\lambda \sigma}{2})$$

The first step is a Landweber iteration, used for solving least squares problems. Notice that

$$A^T(y - Ax_{k-1}) = -\frac{1}{2}\nabla_x \|y - Ax\|_2^2 \Big|_{x=x_k} \tag{17}$$

Landweber iteration is fundamentally a gradient descent step with stepsize σ. In [13, 14], the gradient descent step is generalized from linear to non-linear functions.

$$b = x_{k-1} - \frac{\sigma}{2}\nabla_x \|y - f(x)\|_2^2 \Big|_{x=x_k} \tag{18}$$

The soft thresholding step remains as it is, albeit with a different threshold.

$$x_k = signum(b)\max(0, |b| - \tau) \tag{19}$$

The problem is in deciding the step size and the threshold value. For linear problems these could be evaluated easily. But for non-linear problems they need to be fixed empirically. In theory, the step size can be evaluated from the Lipschitz constant, but in practice this value is too pessimistic which decelerates the convergence of the algorithm. To get good performance, these parameters need to be tuned.

### 3.3. Proposed Algorithm

The objective is to solve (9). The generic form for the analysis prior sparse recovery problem from non-linear measurements is given by:

$$\|y - f(x)\|_2^2 + \lambda \|\Psi x\|_1 \tag{20}$$

For solving is problem, we follow an approach similar to that of the synthesis prior algorithm. As can be seen from the linear problem (12), (15) and the non-linear synthesis prior problem (18), the first step is a gradient descent step. For linear measurements, it was the simple Landweber iteration (15); for the current scenario it will be a non-linear version of it, given by (18) - we repeat it for the sake of completion.

$$b = x_{k-1} - \frac{\sigma}{2}\nabla_x \|y - f(x)\|_2^2 \Big|_{x=x_k}$$

This allows (20) to be written as follows in every iteration:

$$\|b - x\|_2^2 + \lambda \|\Psi x\|_1 \tag{21}$$

This is exactly the same as (15). Hence the updates for solving (21) will be the same as (15). They are:

$$z_k = (\frac{2a}{\lambda}D^{-1} + cI)^{-1}(cz_{k-1} + \Psi(b - \Psi^T z_{k-1})) \tag{22a}$$

where $D = diag(|\Psi b|^{-1})$

$$x_k = b - \Psi^T z_k \tag{22b}$$

## 4. Experimental Results

### 4.1. Benchmarking Experiments

In the first set of experiments we test our algorithm:

1. We empirically study the convergence of the proposed algorithm
2. We empirically study the success rate of the proposed algorithm

To study the convergence, we consider three measurement functions - Linear ($f(x) = Ax$), Exponential $f(x) = \exp(Ax)$) and Logarithmic ($f(x) = \log(Ax)$). The projection matrix A is an i.i.d Gaussian matrix of size 40 X 100. The number of zeroes in Ψx is 10. For all cases Ψ is the complex dualtree wavelet.

We study the convergence of the analysis prior sparse recovery algorithm for linear measurement function. Vis-a-vis we study the convergence of our algorithm for the linear, exponential and logarithmic functions.

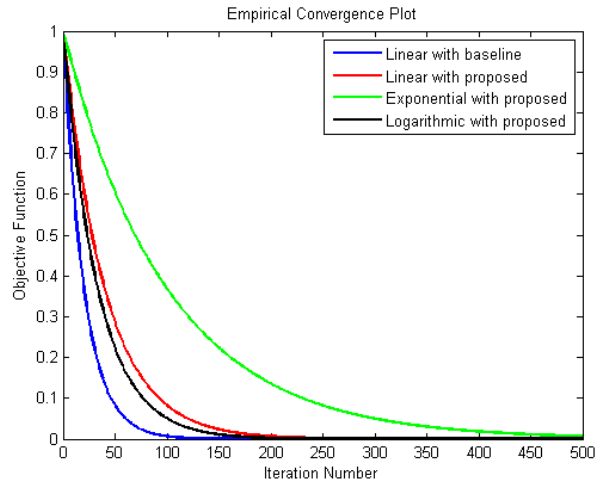

**Figure 1.** Empirical Convergence Plot: Comparison of Baseline (Algorithm for recovering sparse signals from linear measurements) vs Proposed

We find that our algorithm converges, albeit slower compared to the baseline. This is understandable; the baseline has an analytically derived step-size; whereas for ours, it needs to be tuned.

Next we test the empirical success rate of our method compared to the baseline. The problem setup remains similar as before. But the number of non-zero elements in Ψx is varied. For each number of non-zero elements, A is generated 1000 times and for each such A, the normalized mean squared error between the actual and the recovered signal is computed. If the error is smaller than $10^{-3}$, the trial is considered successful. The empirical convergence rate is the plot of success rate versus the number of non-zeroes in Ψx.

Empirical success rate is a popular technique to compare sparse recovery techniques. In this case, we do not have an algorithm to benchmark against. So we compare against the well known algorithm for analysis prior sparse recovery for linear measurements. We find that, even for linear measurements our proposed algorithm gives a slightly better success rate compared to the baseline. Results for exponential function are even better. Unfortunately success rate for logarithmic function is poor.

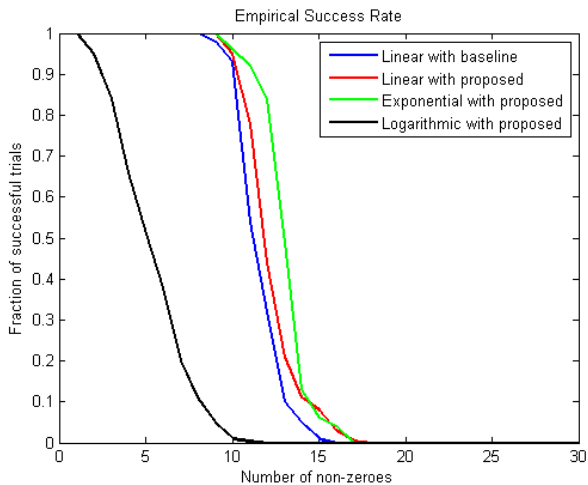

**Figure 2.** Empirical Success Rate: Comparison of Baseline (Algorithm for recovering sparse signals from linear measurements) vs Proposed

### 4.2. Experiments on T1 Map Estimation

The only prior work in this area is [9]. A matrix factorization based model was used for solving estimating quantitative maps. In the said study, the recovery was performed not on the K-space but on the images – it assumed a Gaussian noise model on the image. However, this is erroneous since it is well known that the noise in the image space is Rician and not Gaussian and hence cannot be removed by simple $l_2$-norm minimization; that is why most modern techniques propose estimating directly from the K-space [17]. Although the previous technique was used for T2 maps, it is general enough to be applied for T1 map estimation as well.

In this work, we propose to sub-sample the K-space for both PD and T1 estimation. This has not been attempted in [9]; there in, it is assumed that the image is generated from the fully sampled K-space. Our objective is to have a scan time comparable to that of an ordinary scan, i.e. the total number of K-space samples for the corresponding PD and T1 scans should add up to a complete K-space scan.

We follow an experimental protocol similar to the previous work [9]. We assume a saturation recovery spin echo sequence; there are other fast mapping techniques based on pulse engineering and sequence designing, but we are not discussing them in this study.

We show simulation results; the experiments need to be simulated because T1 maps cannot be directly measured, they are always estimated from images. Our proposal is a new (direct) technique for estimating these maps. Therefore we cannot assume previous techniques to be yield absolutely correct groundtruth maps. The only option is to simulate the groundtruth proton density and T1 maps and generate MR images from these. Our proposed technique and previous ones use the generated MR images to estimate the proton density and T1 maps; the estimated maps are compared with the original to test the efficacy of the techniques. Fig. 3. shows the original proton density (PD) and T1 maps. These are normalized for display.

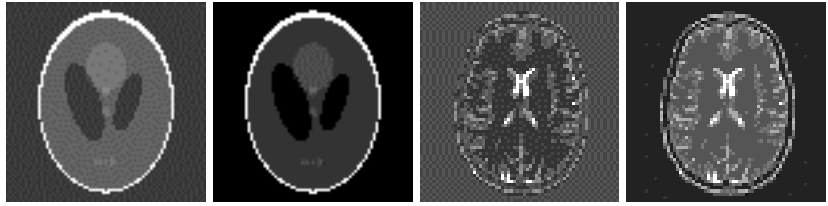

**Figure 3.** Left to Right – Shepp-Logan PD , Shepp-Logan T1, Brain PD and Brain T1

To estimate the PD, the TR is kept such that the effects of T1 is nullified. This is achieved by putting TR to be relatively high. To get the T1 map, the TR is of the same order as the range of T1; we keep TR = average of T1. The results are shown in the following Fig. 4. For sub-sampling the K-space we simulated variable density random sampling.

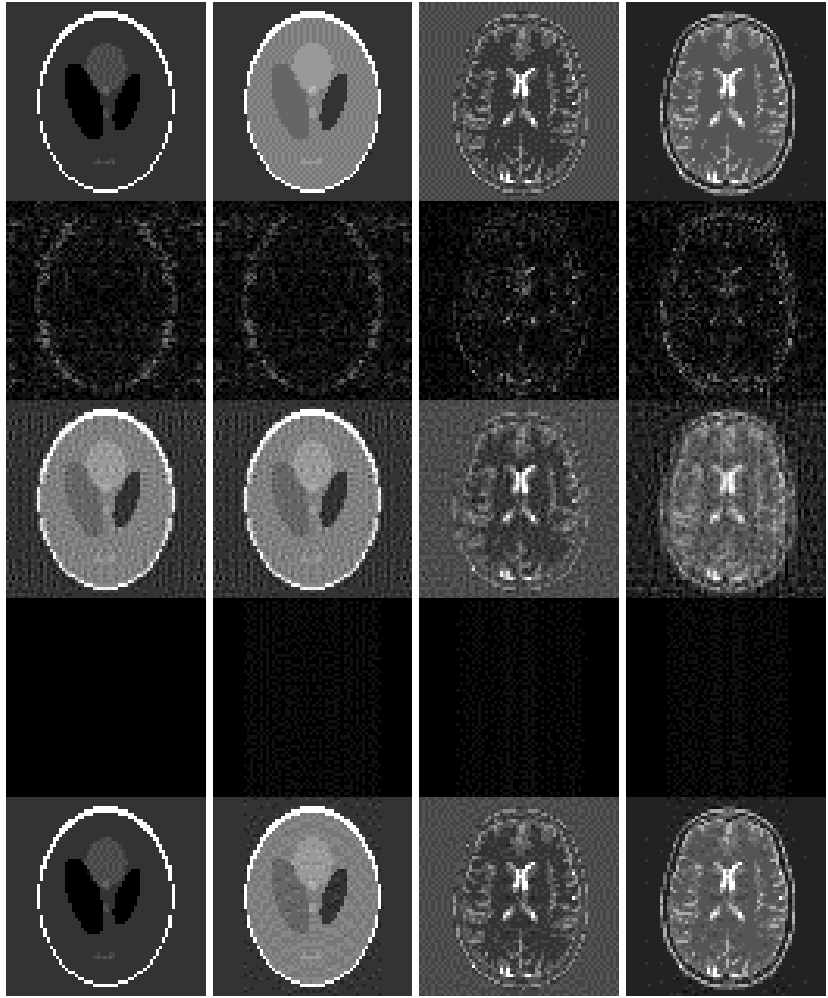

**Figure 4.** 1$^{st}$ Row – Groundtruth, 2$^{nd}$ Row – Difference images from [9], 3$^{rd}$ Row – Reconstruction from [9], 4$^{th}$ Row – Difference using proposed method and 5$^{th}$ Row – Reconstruction using proposed method. Left to Right – Shepp-Logan PD, Shepp-Logan T1, Brain PD and Brain T1.

For our method, it is assumed that the image is sparse in finite difference domain leading to a small total variation - this applies both to the PD as well as the T1 map. The PD is reconstructed from 30% K-space samples by solving (6). Once we obtain the PD, we compute the T1 map from the partially sampled (70%) K-space corresponding to the second image using our proposed technique. The results are shown in Fig. 4. We have also compared the results from the previous [9] technique for T1 map estimation. It can be seen that the recovery from the prior technique is highly noisy; there are a lot of reconstruction artifacts. For our method the PD recovery (in both cases) is almost perfect, but there are some artifacts in solving the non-linear problem for T1 recovery. The artifact is more pronounced in the Shepp-Logan phantom because it has a regular geometrical structure.

For numerical evaluation we show the errors from the previous technique [9] and the proposed technique in Table 1. In the first column, the percentage of K-space sampling corresponding to the PD and T1 echoes are shown. We find that as the sampling ratio for PD increases, the reconstruction error for PD decreases. For T1, best reconstruction is obtained when the PD is error is reasonably low while the sampling ratio for the T1 echo is high - this is achieved as 30% K-space sampling for PD and 70% K-space sampling for T1. When the sampling ratio for PD is lower, the PD estimate is poor; since the PD is required for estimating the T1 map, the estimate for the T1 map is poor as well. When the sampling ratio for PD is higher, the sampling ratio for the T1 map is low, this naturally hampers the reconstruction of the T1 map.

Table 1. Normalized Mean Squared Error for PD and T2 Recovery.

| Sampling Ratios | PD Error – Previous [9] | T1 Error - Previous [9] | PD Error – Proposed | T1 Error - Proposed |
|---|---|---|---|---|
| 20%PD 80%T2 | 0.112 | 0.231 | 0.056 | 0.093 |
| 30%PD 70%T2 | 0.067 | 0.126 | 0.024 | 0.025 |
| 40%PD 60%T2 | 0.030 | 0.259 | 0.011 | 0.114 |
| 50%PD 50%T2 | 0.009 | 0.338 | 0.006 | 0.226 |

## 5. Conclusion

The problem we address in this work is fast acquisition for T1 maps. Via simulation results we show that it is possible to speed up the acquisition time to an extent where Proton Density (PD) and T1 maps can be estimated from scans which roughly consume the same amount of time as a standard MR image.

We partially sample the K-space such that the scan for the PD and T1 add up to the total of one full K-space. Recovering the PD is straight-forward using compressed sensing techniques. But recovering the T1 maps from partial K-space measurements is challenging; this is because the relationship between the K-space scan and the T1 map is non-linear. This demands solution of a non-linear sparse recovery problem. Literature on such problems is parsimonious. In this work, we derive an algorithm to solve this problem.

We did simulation studies to evaluate the efficacy of our method. These studies show promise. It remains to be seen in the future how the technique performs on real data.